\documentclass{article}

\usepackage{PRIMEarxiv}

\usepackage[utf8]{inputenc} 
\usepackage[T1]{fontenc}    
\usepackage{hyperref}       
\usepackage{url}            
\usepackage{booktabs}       
\usepackage{amsfonts}       
\usepackage{nicefrac}       
\usepackage{microtype}      
\usepackage{lipsum}
\usepackage{fancyhdr}       
\usepackage{graphicx}       
\graphicspath{{media/}}     

\title{Bringing Comparative Cognition To Computers
\thanks{\textit{\underline{Citation}}: 
{Voudouris, K., Cheke, L. G., \& Schulz, E. (2025) \textit{Bringing Comparative Cognition To Computers.} arXiv Pre-print.}} 
}

\author{
  Konstantinos Voudouris \\
  Human-Centered AI, Helmholtz Munich \\
  Centre for the Future of Intelligence, University of Cambridge \\
  \texttt{k.voudouris@srcf.net} \\
   \And
  Lucy G. Cheke \\
  Department of Psychology, University of Cambridge \\
Centre for the Future of Intelligence, University of Cambridge \\
  \texttt{lgc23@cam.ac.uk} \\
    \And
  Eric Schulz \\
  Human-Centered AI, Helmholtz Munich \\
  \texttt{eric.schulz@helmholtz-munich.de} \\
}

\begin{document}
\maketitle

\begin{abstract}
Researchers are increasingly subjecting artificial intelligence systems to psychological testing. But to rigorously compare their cognitive capacities with humans and other animals, we must avoid both over- and under-stating our similarities and differences. By embracing a comparative approach, we can integrate AI cognition research into the broader cognitive sciences.
\end{abstract}

\section{Introduction}

Artificial intelligence (AI) systems, from large language models (LLMs) to reinforcement learning agents, now exhibit behaviours once assumed to be exclusive to humans and other animals. As such, researchers are increasingly probing these systems using psychological methods, asking questions about how they explore new environments, make decisions in risky conditions, and reason about their own uncertainty \cite{hagendorff2023machine}. This work appears to be driven by two motivations: better characterising what AI can and cannot do so that we can improve it and use it safely; and the tantalising proposition that AI constitutes a new class of cognitive system worthy of serious scientific attention, not only to learn more about how they work but to better understand our own cognition \cite{simon1980cognitive}.

But applying methods designed for human cognitive psychology to test AI risks both under- and over-attributing cognitive capacities to them - because those tests may be ill-designed for these non-human subjects. Comparative cognition - the study of non-human animal behaviour - has grappled with similar challenges for decades. By adopting its methods, AI research could avoid pitfalls, join the cognitive sciences, and clarify the nature of cognition itself.

\section{The Cognition Thesis}

Cognition is notoriously difficult to define. Whether it is defined as computations over representations, as information-processing, or as an emergent property of a complex system, most agree that it involves mechanisms like learning, memory, and decision-making that drive flexible and adaptive behaviour \cite{bayne2019cognition}. The archetypal example of a system that exhibits these sorts of cognitive capacities is of course humans, and in both scientific and non-scientific contexts, our cognition is used as a reference against which to compare other systems. Indeed, throughout the history of psychology and the cognitive sciences, researchers have sought evidence of similar human-like capacities in other systems. In so doing, they have stretched and adapted the definition of cognition, often in an effort to emancipate it from its inherent anthropocentrism, such that we can understand our behaviour in the context of a tapestry of complex behavioural systems. One way of interpreting this practice is as an endorsement of what we call the cognition thesis - that cognition can emerge from many structurally distinct systems, if they are arranged appropriately. It is the task of cognitive science to clarify the nature of that arrangement.

Earliest in this line of thinking were those who sought evidence for cognition in non-human animals, particularly primates, birds, and cetaceans like dolphins and whales \cite{shettleworth2009cognition}. More recently, cognition and related cognitive terminology has been applied to invertebrates, most notably bees, ants, and cephalopods such as octopus and cuttlefish \cite{perry2013invertebrate}. Recent years have also witnessed an explosion in the study of cognition in non-neural systems, including bacteria \cite{lyon2015cognitive}, plants \cite{segundo2022consciousness}, and protists like the slime mould \cite{reid2023thoughts}. For subscribers to the cognition thesis, an intriguing prospect is emerging as they observe the rise of sophisticated artificial intelligence. Could cognition emerge not only from living biological systems, but from computers too? This idea has gained significant traction with the rise of large language models (LLMs), those systems that can process text, and oftentimes, images and sounds too. Indeed, innumerable claims have been made about the human-like behaviour of these models, often applying the same vocabulary used to describe human behaviour to describe LLM behaviour \cite{hagendorff2023machine, shevlin2019apply}.

But claims of AI cognition remain controversial, with debates over the validity of the experiments and the strength of the available data. This bears a remarkable similarity to the debates that surrounded non-human animal cognition in its formative years in the early twentieth century, in response to which animal psychologists developed numerous tools to evidence and justify their claims \cite{boakes1984darwin}. We propose that cognitive scientists interested in AI cognition should learn lessons from contemporary comparative cognition to make progress.

\section{Lessons From Comparative Cognition}
If we are to make progress on studying cognition in computers, we must confront a fundamental challenge: how to compare our behaviour with that of systems profoundly different from humans, so that we can draw out our commonalities as well as our differences. Human-centric psychological tests often fail to accommodate the unique architectures and experiences of non-human subjects. This means that we may fail to attribute cognitive capacities to them when they have them, as well as prematurely attributing cognitive capacities where they lack them. Comparative cognition - the study of animal minds - has grappled with this problem for over a century, developing tools and frameworks that AI researchers can adapt to avoid similar pitfalls.

\subsection{Avoiding Under-Attribution}

Comparative cognition abounds with examples of experimental designs that initially obscured true cognitive capacities. Take the case of domestic dogs and object permanence - the ability to track hidden objects, a key feature of human visual cognition \cite{piaget1923origins}. Early studies concluded that dogs lacked this capacity because they performed poorly on the invisible displacement task, where a reward is hidden in a movable container which is then moved to a new location behind an occluder. The reward is hidden there and the movable container is shown to be empty. Dogs fail to robustly locate the reward in its new location in these scenarios, suggesting they cannot track its movement while it is occluded. However, later research revealed that the task itself was the problem: dogs get distracted by the container and track that instead \cite{muller2014use}. Subsequent studies using several independent measures of object permanence now suggest that dogs have object permanence \cite{zentall2016now}.

Cognitive scientists investigating AI systems should avoid the pitfall of mistakenly under-attributing cognitive capacities to these systems by paying attention to the validity of the tests they use. For instance, LLMs appear to struggle with arithmetic involving large numbers, inviting the conclusion that they lack a fundamental component of human cognition, our ability to manipulate numbers. However, their apparent arithmetical inability often stems from tokenization - the way LLMs process text. LLMs do not see the text the way we do, as individual characters collected into words on a page. Instead, they combine characters together into tokens that are represented as atomic symbols - the exact characters that go into these tokens are learnt from data, with a preference for grouping together commonly co-occurring characters. This means that commonly occurring numbers like ‘100’ and ‘99’ are seen as single units. This inevitably leads to problems when the model has to compute sums like ‘100100100 + 999999’. Indeed, when this problem of tokenization is minimized, the performance of LLMs greatly increases \cite{singh2024tokenization}.

\subsection{Avoiding Over-Attribution}

Of course, correcting the potential under-attribution of cognitive capacities risks over-attribution. The infamous case of Clever Hans, the horse in the 1900s who appeared to solve arithmetic problems, illustrates this danger \cite{beran2012did}. While many at the time were astonished by his numerical acumen, it later transpired that Hans’s owner was giving him subtle, and unconscious, cues to the right answer, creating the illusion of mathematical reasoning. This parable remains a cornerstone of comparative research, reminding researchers to rigorously rule out alternative explanations for seemingly sophisticated behaviour.

AI research has its own Clever Hans moments. Deep neural networks, for instance, excel at identifying common objects from visual images, discriminating between millions of objects with an accuracy that sometimes surpasses that of humans. However, given images that are imperceptibly different to a human observer, their accuracy unexpectedly collapses, leaving in tatters the claim that these models have a human-like object recognition capacity. Instead, these models mostly rely only on superficial features, such as textures or patterns, revealing brittle, non-human-like forms of object recognition \cite{ilyas2019adversarial}. Without scrutinising these tests to determine if a Clever Hans effect could be at play, we risk overstating how similar AI systems are to humans and other animals in terms of their cognitive capacities.

\section{Towards a Comparative Science of AI}

The rise of AI demands a new comparative science - one that borrows methods from animal cognition to study machines. In particular, we must rigorously evaluate the experiments we design to ensure that we do not overstate nor underplay the cognitive capacities of these AI systems. To do so, individual researchers as well as the wider scientific community must scrutinise experiments to determine if they adequately control for alternative explanations and take into account the constraints of the system under study. This demands a truly comparative approach to cognition—one that explicitly evaluates both the similarities and differences between AI systems, humans, and other animals in terms of their cognitive capacities. By embracing this challenge, we can make progress on the questions that unite all cognitive scientists: What is cognition, and where does it come from?

\bibliographystyle{unsrt}  
\bibliography{references}

\end{document}